\documentclass[11pt,a4paper]{article}
\usepackage[utf8]{inputenc}
\usepackage[T1]{fontenc}
\usepackage{lmodern}
\usepackage[a4paper,margin=20mm]{geometry}
\usepackage{microtype}
\usepackage[table,dvipsnames]{xcolor}
\usepackage{graphicx}
\usepackage{amsmath,amssymb}
\usepackage{booktabs,array,multirow}
\usepackage[numbers,sort&compress]{natbib}
\usepackage{etoolbox,xspace}
\usepackage{placeins,float,multicol}
\usepackage{tikz}
\usepackage[font=small,labelfont=bf,labelsep=period]{caption}
\usepackage{subcaption}
\usepackage{titlesec}
\titleformat{\section}{\large\sffamily\bfseries}{\thesection}{0.7em}{}
\titleformat{\subsection}{\normalsize\sffamily\bfseries}{\thesubsection}{0.7em}{}
\titleformat{\subsubsection}{\normalsize\sffamily\bfseries}{\thesubsubsection}{0.7em}{}
\titlespacing*{\section}{0pt}{2.2ex plus 1ex minus .2ex}{1.1ex plus .2ex}
\titlespacing*{\subsection}{0pt}{1.8ex plus 1ex minus .2ex}{0.8ex plus .2ex}
\definecolor{linkblue}{HTML}{245B80}
\usepackage[colorlinks,allcolors=linkblue,breaklinks]{hyperref}
\hypersetup{
  pdftitle={BOT-Scale: A Traceable Evaluation Methodology for Robotic Manipulation},
  pdfauthor={Yunfei Xue, Kemi Chen, Weiling Chen, Wei Gao, Tiesong Zhao, Zheng-Jun Zha},
  pdfsubject={Robotic manipulation evaluation},
  pdfkeywords={robotic manipulation, behavioral assessment, traceable evaluation}
}
\makeatletter
\newenvironment{figuretablepair}{\@float{figure}[tbp]}{\end@float}
\makeatother

\newlength{\correspondencecontentwidth}
\definecolor{tabBlueTint}{HTML}{F6F9FD}
\definecolor{tabPeachTint}{HTML}{FCFAF6}
\definecolor{tabGreenTint}{HTML}{F5FAF7}
\definecolor{tabLavTint}{HTML}{F8F7FC}
\definecolor{tabBlueMid}{HTML}{EBF0FA}
\definecolor{tabBlueBorder}{HTML}{2F67B1}
\definecolor{tabBlueDark}{HTML}{234A78}
\definecolor{frameFirst}{HTML}{F1F6FC}
\definecolor{frameSecond}{HTML}{D7E5F4}
\definecolor{frameThird}{HTML}{A7C4E2}
\definecolor{frameLast}{HTML}{476D94}
\definecolor{tabCoralDeep}{HTML}{A24E24}
\definecolor{tabGreenDeep}{HTML}{25794E}
\definecolor{tabLavDeep}{HTML}{5C4FA3}
\definecolor{tabGreen}{HTML}{2E9D67}
\definecolor{tabRed}{HTML}{C85A5A}

\definecolor{tabRankBlueBest}{HTML}{AFC8E5}
\definecolor{tabRankBlueSecond}{HTML}{ECF3FA}
\definecolor{tabRankGrayBest}{HTML}{C5C5C5}
\definecolor{tabRankGraySecond}{HTML}{F2F2F2}
\definecolor{tabOursGold}{HTML}{FDF1D0}
\definecolor{tabOursGoldSecond}{HTML}{FEF8E8}

\title{\sffamily\bfseries BOT-Scale: A Traceable Evaluation Methodology for Robotic Manipulation}
\author{\normalsize\begin{tabular}{ccc}
Yunfei Xue & Kemi Chen & Weiling Chen \\[4pt]
Wei Gao & Tiesong Zhao & Zheng-Jun Zha
\end{tabular}}
\date{}
\begin{document}

\maketitle

\begin{abstract}
A successful manipulation rollout can still involve a mid-task object drop or a placement deviation. Existing evaluators reduce such deviations to a single score or attach explanations computed separately; the score itself explains neither its basis nor what to fix.
We show that a manipulation score becomes traceable when defined as the sum of outcome-conditioned penalties over retained behavioral-factor records. Explanation and score share the same computation, so each factor-level deduction names an inspectable deviation.
BOT-Scale applies this principle by adapting behavioral assessment scales from medicine. It organizes twelve candidate factors into four assessment layers; each factor is tied to observable evidence and an explicit decision criterion.
Reference calibration fixes a penalty per factor, and a closed-form rule aggregates the factor records into a score with per-factor contributions.
We developed BOTScorer, an automatic scoring system based on the seven factors retained by data-driven screening, to validate this methodology.
We provide BOT-Events, a dataset with eight-rater quality scores and factor-event annotations, alongside the traceable, interpretable score.
The code will be released publicly.
\end{abstract}

\section{Introduction}\label{sec:intro}

Robotic manipulation is advancing rapidly with Vision-Language-Action (VLA) models~\cite{kim2024openvla,black2024pi0,intelligence2025pi05}. Sustaining this progress requires evaluation that attributes performance to behaviors a practitioner can fix. We show that a manipulation score becomes traceable when defined as the sum of outcome-conditioned penalties over retained behavioral-factor records: each score gap then decomposes into named, inspectable deviations that make its basis explicit.

Automated evaluation of robotic manipulation now extends beyond success rates: RoboJev~\cite{chen2026robojev} infers success probabilities from rollout video; PRM-as-a-Judge 1.5 (PRM-SQS)~\cite{liu2026prmjudge} derives fine-grained process metrics from dense progress curves; RoboMeter~\cite{liang2026robometer} learns rewards from comparisons of expert and suboptimal trajectories; and Eval-Actions/AutoEval~\cite{liu2026evalactions} performs criteria-level expert grading with diagnostic explanations.

These evaluators share three limitations. \textbf{First,} a single score hides its behavioral basis because it does not specify which deviations occurred. It cannot distinguish a clean success from a rollout that succeeds despite a force-impedance control failure or a transient end-effector positioning failure, both of which directly affect stability, efficiency, and safety. \textbf{Second,} when explanations are available, they are computed separately from the score. That score cannot be recomputed from the supporting records. \textbf{Third,} no evaluator exposes per-factor contributions, leaving a score gap untraceable to named, fixable deviations. Comparing manipulation quality therefore requires both a score and process evidence explicitly linked to its computation.

\begin{figure}[!t]
\centering
\includegraphics[width=\linewidth]{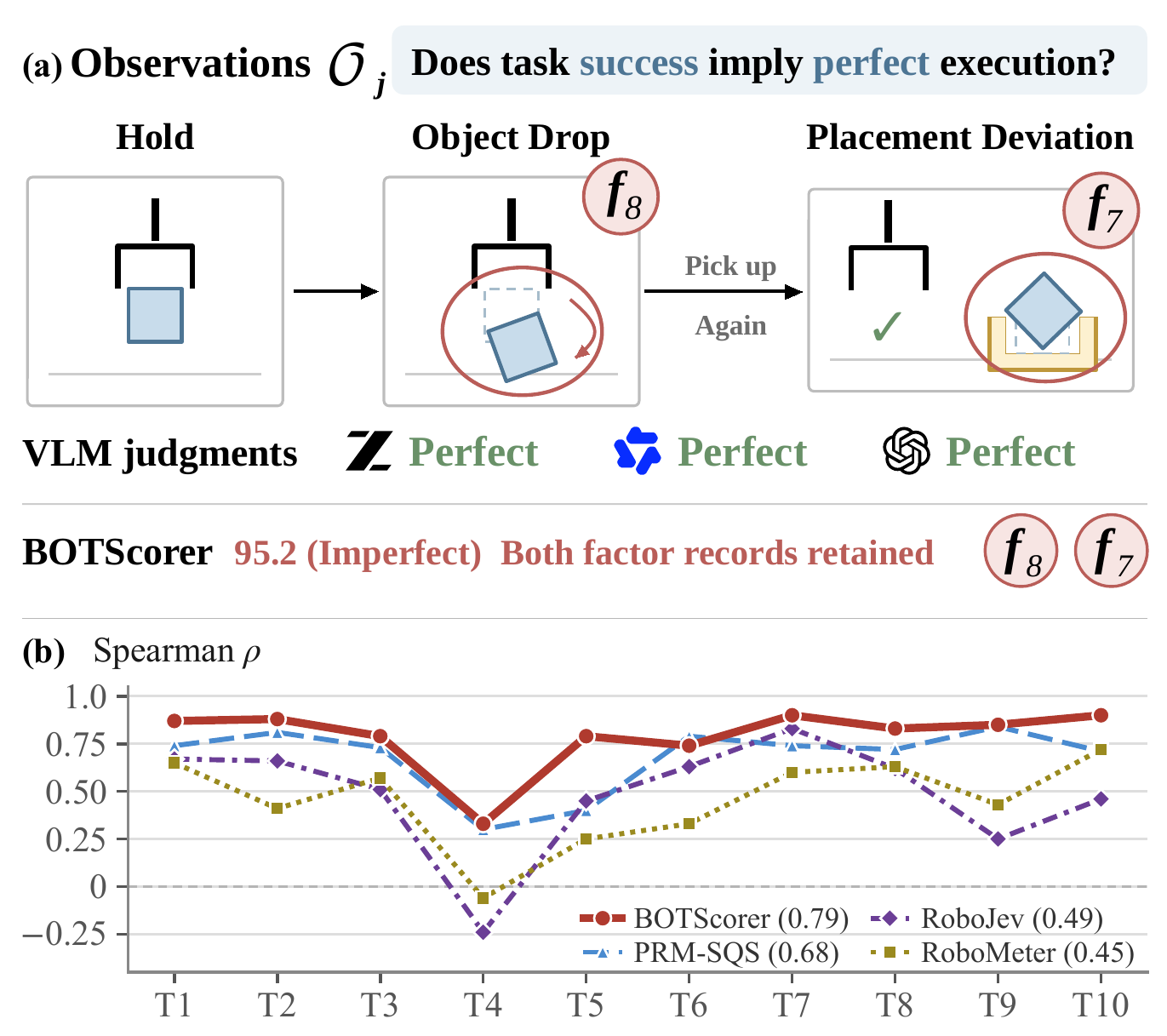}
\caption{(a) GLM-high, Qwen3-VL, and
GPT-6.1-sol (light) judge a successful rollout
perfect, while BOTScorer records f8/f7 and scores 95.2.
(b) Per-task Spearman correlations with the eight-rater
Mean Opinion Score (MOS) on 800 videos across ten tasks.}
\label{fig:task-relative-assessment}
\end{figure}

BOT-Scale addresses these limitations (Fig.~\ref{fig:task-relative-assessment}). Drawing on behavioral assessment scales in medicine, it records the behavioral basis of a score: each deviation becomes a behavioral factor, an assessment item with observable evidence and an explicit decision criterion. Twelve candidates, organized into four assessment layers (Table~\ref{tab:scale-correspondence}), are screened to seven. The score is then defined as the sum of outcome-conditioned penalties over the retained records (Eqs.~\ref{eq:rollout-deduction}--\ref{eq:score-contributions}). Explanation and score share the same computation, making every score recomputable. The closed-form rule yields per-factor contributions that decompose each score gap into named, fixable deviations.

BOTScorer is the automatic scoring system that implements this methodology; BOT-Events supplies the data for its calibration and validation. BOTScorer uses a discriminative detector to recognize deviation events of the seven retained factors in execution video. Reference calibration and a closed-form rule aggregate these records into a recomputable score with per-factor contributions. These scores agree more closely with human ratings than learned aggregators that use the same records and are trained on those ratings (Sec.~\ref{sec:mos-consistency}). BOT-Events contains 800 rollouts with eight-rater quality scores and 2,394 factor-event annotations, used to validate BOT-Scale and to build BOTScorer.

Our main contributions are:
\begin{itemize}
\item \textbf{BOT-Scale.} We define quality as the sum of outcome-conditioned penalties over retained behavioral-factor records. The explanation is the computation itself. Adapting the item-generation, evidence-and-criterion, domain-grouping, and item-analysis steps of clinical behavioral assessment scales, we specify twelve candidate factors in four assessment layers and retain seven through screening. Their pairwise correlations remain low, so the additive decomposition does not charge the same deviation twice.
\item \textbf{BOTScorer.} A discriminative detector recognizes factor events in execution video, reaching 90.38\% event accuracy, 2.1 points below generative label decoding on the same backbone, at $1.94\times$ the speed. Among successful executions, scores computed from these records reach a mean per-task Spearman $\rho$ of 0.59, compared with at most 0.22 for three released evaluators; agreement with human quality judgments over all videos is 0.79. The same fixed penalties also distinguish two VLA checkpoints differing by only 0.2 points in Success Rate (SR).
\item \textbf{BOT-Events.} We provide 800 rollouts with 6,400 human quality ratings. The mean task-wise Intraclass Correlation Coefficient (ICC), computed as ICC(2,8), is 0.951; eight-rater split-half reliability is 0.959. The dataset also includes 2,394 factor-event annotations and the calibrated penalties needed to recompute every reported score.
\end{itemize}

\section{Related Work}\label{sec:related}

\subsection{Manipulation Benchmarks}
LIBERO~\cite{liu2023libero} reports performance through success rates. LIBERO-Elegant further evaluates execution constraints~\cite{mao2026beyondsuccess}. RoboEval adds efficiency, coordination, and safety metrics~\cite{wang2025roboeval}, whereas MetaFine distinguishes among understanding, perception, and controlled behavior~\cite{xu2026metafine}. These metrics appear alongside scores, but they are not integrated into the computation of those scores.

\subsection{Evaluation of Robotic Manipulation}
\textbf{Single-score evaluators} summarize an execution in one number: RoboJev predicts success probabilities from instructions and sampled video frames~\cite{chen2026robojev}; Valle et al.~\cite{valle2026vlaquality} compare quality and uncertainty metrics against expert success judgments; RoboMeter learns rewards from expert frame-level progress and within-task trajectory preferences, including suboptimal and failed executions~\cite{liang2026robometer}; PRM-as-a-Judge 1.5 derives failure-side progress, recovery, and success-side quality metrics from dense progress curves~\cite{liu2026prmjudge}. These evaluators retain no behavioral records, so scores cannot be recomputed from the deviations that produced them.

\textbf{Explanation-oriented evaluators} attach diagnoses to scores. Eval-Actions combines criteria-level expert grading, calibrated physical indicators, and diagnostic annotations over 13{,}000+ real-robot episodes and 150+ tasks~\cite{liu2026evalactions}. It uses expert rubrics, whereas BOT-Scale assesses task-relative deviations screened on reference data. AutoEval predicts quality and generates diagnostic explanations from video and kinematic summaries~\cite{liu2026evalactions}. The explanations are computed separately from the score. The score therefore cannot be recomputed from the records supporting those explanations, and per-factor contributions are not exposed. BOT-Scale uses the behavioral records themselves as scoring terms, so explanation and score share one computation.

\begin{table*}[!t]
\caption{How human behavioral assessment guided the twelve candidate
factors. For each factor: the assessment item its design references
among the cited instruments, and the template element it borrows. The
instruments contribute the construction procedure and the item format;
factor content remains task-derived. Full factor observations and
criteria are in Supplementary Table~A14.}
\label{tab:scale-correspondence}
\centering
\fontsize{8.2}{9.8}\selectfont
\setlength{\tabcolsep}{1.5pt}
\setlength{\correspondencecontentwidth}{\dimexpr\linewidth-8\tabcolsep\relax}
\renewcommand{\arraystretch}{1.25}
\setlength{\extrarowheight}{0pt}
\begin{tabular}{>{\centering\arraybackslash}p{0.024\correspondencecontentwidth}
 >{\raggedright\arraybackslash}p{0.220\correspondencecontentwidth}
 >{\raggedright\arraybackslash}p{0.444\correspondencecontentwidth}
 >{\raggedright\arraybackslash}p{0.312\correspondencecontentwidth}}
\toprule
\rowcolor{tabBlueMid}
\textbf{ID} & \textbf{Factor} & \textbf{Referenced assessment item} & \textbf{Borrowed template element} \\
\midrule
\rowcolor{tabBlueTint} & \multicolumn{3}{l}{\textcolor{tabBlueDark}{\textbf{Perception and understanding}}} \\
\rowcolor{tabBlueTint}
f1 & Progress-tracking failure & economy-of-motion items in technical-skill rating~\cite{martin1997osats,miao2026behavior} & progress features: mid-task stalls, retries \\
\rowcolor{tabBlueTint}
f2 & Temporal-order confusion & task-step sequencing in ADL-based assessment~\cite{katz1963adl} & order judged against required subtask states \\
\rowcolor{tabBlueTint}
f3 & Physical-state misjudgment & compositional action--state rules in action quality assessment~\cite{okamoto2024nsaqa} & action contradicting the physical relation \\
\midrule
\rowcolor{tabPeachTint} & \multicolumn{3}{l}{\textcolor{tabCoralDeep}{\textbf{Planning and logic}}} \\
\rowcolor{tabPeachTint}
f4 & Primitive-sequence error & flow-of-operation and forward-planning items~\cite{martin1997osats} & prerequisite graph as the decision rule \\
\rowcolor{tabPeachTint}
f5 & Long-horizon forgetting & knowledge-of-the-procedure items (retention across steps)~\cite{martin1997osats} & task goal tracked across subtasks \\
\rowcolor{tabPeachTint}
f6 & Semantic mismatch & command-following and comprehension items (GCS verbal)~\cite{teasdale1974coma} & instruction--action consistency \\
\midrule
\rowcolor{tabGreenTint} & \multicolumn{3}{l}{\textcolor{tabGreenDeep}{\textbf{Execution and control}}} \\
\rowcolor{tabGreenTint}
f7 & End-effector positioning failure & quality-of-result and placement checks~\cite{martin1997osats} & put-down placement versus the target zone \\
\rowcolor{tabGreenTint}
f8 & Force-impedance control failure & respect-for-tissue items (force-appropriate handling)~\cite{martin1997osats} & unintended pose change of a held object \\
\rowcolor{tabGreenTint}
f9 & Unreasonable grasping & instrument-handling items~\cite{martin1997osats} & grasp geometry at contact in one frame \\
\rowcolor{tabGreenTint}
f10 & Bimanual coordination failure & --- (task-derived; no clinical counterpart) & synchronized two-arm action during a task \\
\midrule
\rowcolor{tabLavTint} & \multicolumn{3}{l}{\textcolor{tabLavDeep}{\textbf{Interaction and environment}}} \\
\rowcolor{tabLavTint}
f11 & Disturbance intolerance & response-to-stimulus items (GCS)~\cite{teasdale1974coma} & behavior before vs. after a disturbance \\
\rowcolor{tabLavTint}
f12 & Collision non-compliance & safety-compliance and workspace-respect items~\cite{miao2026behavior} & contact of robot or held object with the scene \\
\bottomrule
\end{tabular}
\end{table*}

\subsection{Behavioral Assessment Scales in Medicine}
Medicine assesses human behavior with standardized scales spanning functional status~\cite{katz1963adl}, consciousness~\cite{teasdale1974coma}, and surgical skill~\cite{martin1997osats}; automated variants ground the grading in video~\cite{miao2026behavior} or explicit compositional rules~\cite{okamoto2024nsaqa}. Outside medicine, HealthBench and Rubrics-as-Rewards score Large Language Model (LLM) outputs against explicit rubric criteria~\cite{arora2025healthbench,gunjal2025rubrics}. BOT-Scale differs in what it aggregates and why (Sec.~\ref{sec:factor-representation}): task-relative deviations whose penalties depend on the recorded outcome, with explanations recomputed from the same records.

\section{Traceable Evaluation Methodology}\label{sec:methodology}

\subsection{Problem Formulation}

Evaluation of robotic manipulation currently relies mainly on success rates and untraceable single scores. These measures are meaningful only when aggregated over large datasets. Because real-world executions are limited in number, these measures alone cannot provide accurate assessment. \textbf{For the constraint:} every deduction must link to named, individually assessable evidence. \textbf{For the objective:} among scores meeting this requirement, we seek the strongest correlation with human quality judgments.

Behavioral deviations are judged relative to task requirements: for rollout $j$, $\mathcal T_j$ collects the requirements (the objective, object and target roles, mandatory action dependencies, and interaction constraints), $\mathcal O_j$ denotes the observations, and $Y_j\in\{0,1\}$ denotes the recorded outcome. Applying a factor's criterion to the observations of a window $w\in\mathcal W_j$ yields a \emph{deviation event}, a concrete instance of one factor. Multiple events of the same factor consolidate into a single rollout record $X_{ji}$, while coexisting factors keep distinct records rather than a single flaw label. A symbol table is in the supplementary material (Sec.~B.1).

We pose scoring as
\begin{equation}\label{eq:problem}
\max_{f}\;\ \rho\bigl(f(\mathcal O_j,Y_j),\ \mathrm{MOS}_j\bigr)
\qquad \text{s.t.}\quad f\in\mathcal F_{\mathrm{trace}},
\end{equation}
where $\rho$ denotes the Spearman correlation across evaluated rollouts, and $\mathcal F_{\mathrm{trace}}$ is the admissible family: scores recomputable in closed form from named factor records, recorded outcomes, and fixed penalties. The scoring rule of Sec.~\ref{sec:scoring} instantiates both requirements. \textbf{For the constraint:} the constraint holds by construction: Eq.~\eqref{eq:score-contributions} decomposes the total deduction exactly. Before the zero floor of Eq.~\eqref{eq:overall-score}, $100-\mathrm{Score}=\sum_{i\in\mathcal I}(C_i^{\mathrm F}+C_i^{\mathrm S})+C_{\mathrm U}$, each term tied to a named record $X_{ji}$ and an explicit penalty. \textbf{For the objective:} the score of Eq.~\eqref{eq:overall-score} is evaluated by the correlation $\rho(\mathrm{Score}_j,\mathrm{MOS}_j)$ between per-rollout scores and the Mean Opinion Score (MOS) of human ratings (Sec.~\ref{sec:mos-consistency}).

\subsection{Human Behavioral Assessment as a Reference}
\label{sec:factor-representation}
Table~\ref{tab:clinical-reference} summarizes the Katz Activities of Daily Living (ADL) index, Glasgow Coma Scale (GCS), and other medical references for Eq.~\eqref{eq:problem}. BOT-Scale borrows their construction procedure and item formats, but derives all factor content from task requirements.

\begingroup
\setlength{\intextsep}{4pt}
\begin{table}[H]
\caption{Medical assessment references. Clinical instruments remain in routine use decades after publication.}
\label{tab:clinical-reference}
\centering\small
\setlength{\tabcolsep}{3pt}
\renewcommand{\arraystretch}{1.00}
\setlength{\extrarowheight}{0pt}
\begin{tabular}{@{}>{\raggedright\arraybackslash}p{0.20\linewidth} >{\raggedright\arraybackslash}p{0.775\linewidth}@{}}
\toprule
\textbf{Reference} & \textbf{Assessment basis and use} \\
\midrule
Katz ADL~\cite{katz1963adl} & Six daily activities (e.g., bathing, dressing), judged by unassisted performance; standard geriatric functional-independence summary. \\
\midrule
GCS~\cite{teasdale1974coma} & Eye/verbal/motor channels with distinct ordered levels; impaired consciousness across raters and settings. \\
\midrule
OSATS~\cite{martin1997osats} & Operative execution: respect for tissue, time/motion, instrument handling, procedure knowledge; OSATS and derivatives used in surgical training/assessment. \\
\midrule
Automated variants & Observed video actions ground grades~\cite{miao2026behavior}; explicit motion rules assess compositional action quality~\cite{okamoto2024nsaqa}. \\
\bottomrule
\end{tabular}
\end{table}
\endgroup

We follow four construction steps. First, generate items from activity failure modes and sort them by observable perspective, as in the GCS channels. Second, name the behavior, evidence, and decision rule; specify applicability, occurrence, and exclusions of similar behaviors. Katz items state activities and qualifying performance levels. Third, group items by what assessors observe, not where behavior originates. Fourth, remove items without variation or discrimination and prune redundancy.

\subsection{Selection of Assessment Factors}
\label{sec:factor-records}
BOT-Scale's assessment item, the \emph{behavioral factor}, names a category of task-relative deviation, with an assessment target, evidence, and a decision criterion. For $K$ candidate factors, we write
\begin{equation}\label{eq:factor-definition}
f_i=\{b_i,\mathcal E_i,\kappa_i\},\qquad i=1,\ldots,K.
\end{equation}
Here $b_i$ is the assessed behavior, $\mathcal E_i$ the required
observations and temporal context, and $\kappa_i$ the criterion under
$\mathcal T_j$. The index $i$ connects each factor's evidence, rollout records, and
scoring contributions. A criterion remains fixed across outcomes and
can apply to different objects or goals under the task requirements; a
judgment retains the factor's identity and its supporting evidence.

Following the scales' domain grouping, we assign factors to four
assessment layers: \emph{perception and understanding} assesses
execution against the task context; \emph{planning and logic} concerns
action dependencies and selection; \emph{execution and control} concerns
physical manipulation; and \emph{interaction and environment} concerns
behavior in the surrounding scene. Like the GCS's observation channels,
the layers organize evidence by the reference frame against which a
deviation is judged. They neither locate errors inside policy modules
nor establish statistical independence; one observed window can support
several layers. Table~\ref{tab:scale-correspondence} records, for each
factor, the referenced item from Sec.~\ref{sec:factor-representation}
and the borrowed evidence format or decision rule, without clinical
content. The layers follow the scales' grouping practice without
copying their domains. New candidates require specified evidence,
criteria, and boundaries; extended criteria are in the supplementary
material (Sec.~B).

Sec.~\ref{sec:scoring} screens the candidates for quantitative scoring.

\section{Automatic Scoring with BOTScorer}\label{sec:scoring}

\subsection{Implementation}\label{sec:implementation}
\begin{figure*}[!t]
\centering
\includegraphics[width=0.971\textwidth]{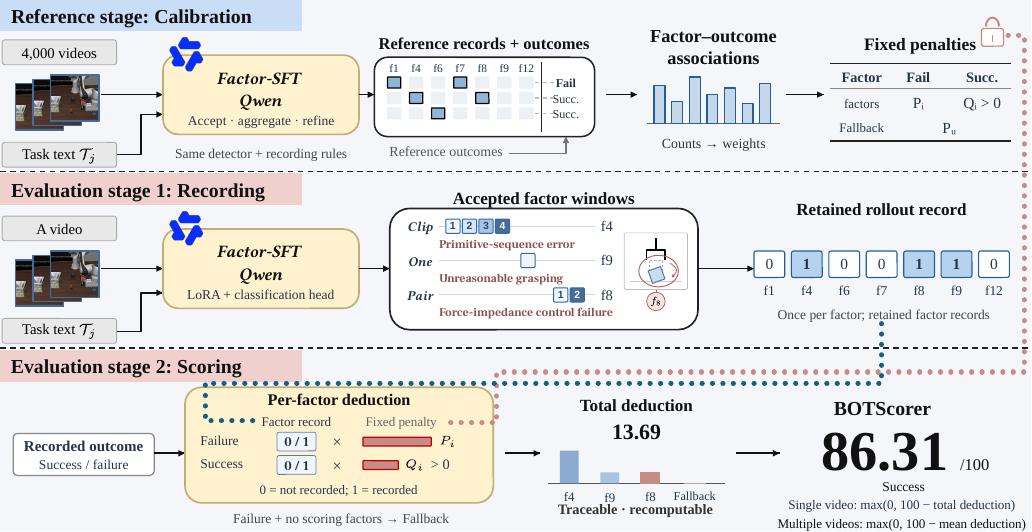}
\caption{Overview of BOTScorer. A shared detector and recording protocol process 4,000 reference videos and one evaluation video. Reference calibration fixes penalties, which combine with factor records and independently recorded outcomes to compute deductions. Positive success-side penalties retain contributions after recovery. Scores equal 100 minus total deductions for one video or mean deductions for multiple videos, floored at zero. Factor and fallback contributions remain traceable and recomputable.}
\label{fig:scoring-overview}
\end{figure*}

BOTScorer implements the BOT-Scale methodology of Sec.~\ref{sec:methodology} in three stages (Fig.~\ref{fig:scoring-overview}): video recognition detects deviation events and produces factor records; reference calibration fixes factor penalties; and a closed-form rule aggregates them into a score with per-factor contributions. The pipeline uses only execution video, task text, and recorded outcomes.

We screen the twelve candidates for quantitative use with two criteria, applying the item-analysis step of Sec.~\ref{sec:factor-representation} to recorded data. Recognition support: the available observations and annotations must support reliable recognition. f10 (no bimanual tasks) and f2, f5, and f11 (no annotated events) are excluded, leaving the detector set $\mathcal D=\{1,3,4,6,7,8,9,12\}$. Calibratability: the reference data must contain enough occurrences to fix a penalty. Within $\mathcal D$, f3 never crosses the recognition threshold on any rollout and is excluded. A scoring factor requires $n_i\ge10$ reference rollouts recording it, with at least one failure association ($n_i^{\mathrm F}\ge1$, quantified in Sec.~\ref{sec:reference-calibration}). Seven factors remain: $\mathcal I=\{1,4,6,7,8,9,12\}$. The methodology defines twelve factors (Table~\ref{tab:scale-correspondence}); this implementation retains the seven it can recognize and calibrate. The decomposition in Eq.~\eqref{eq:score-contributions} remains the same for any retained subset. Annotation statistics are in the supplementary material (Sec.~C).

\begin{figure}[!t]
\centering
\includegraphics[width=0.70\linewidth]{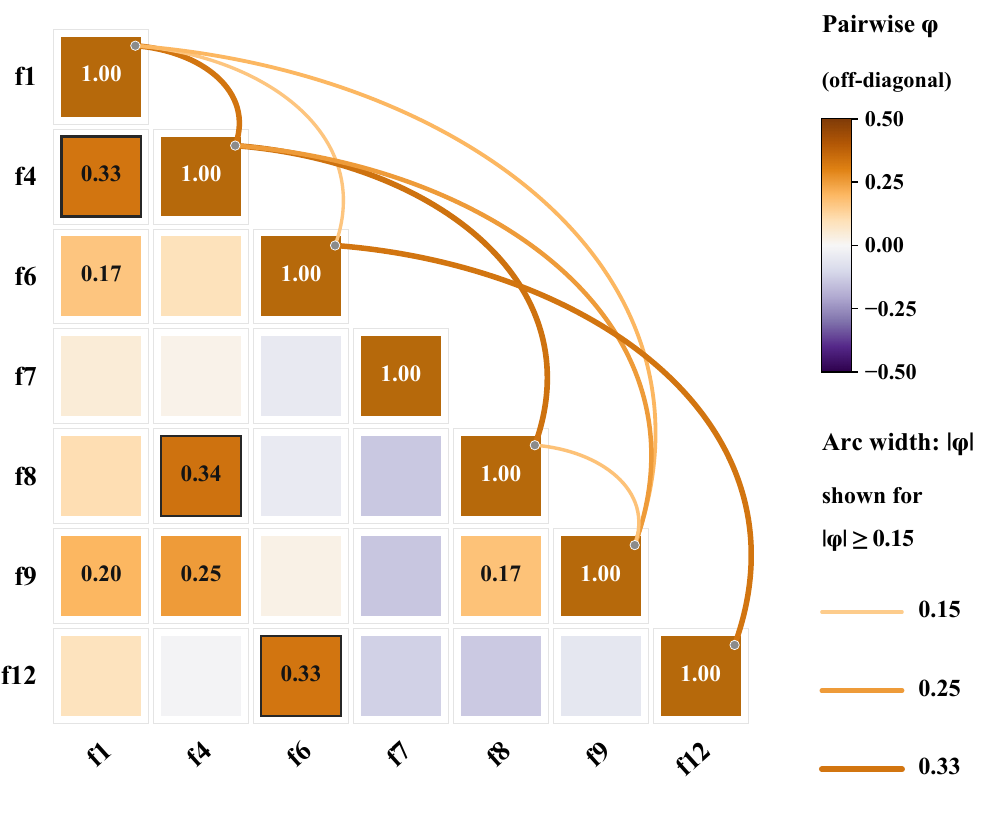}
\caption{Refined factor associations on 4,000 reference rollouts.
Arcs: $|\varphi|\geq0.15$; color and width encode $\varphi$.}
\label{fig:factor-outcome}
\end{figure}

\subsection{Overall Framework}
\label{sec:record-construction}
\label{sec:reference-calibration}
\label{sec:factor-penalties}
\label{sec:score-aggregation}
Factor-SFT-Qwen uses a discriminative head to return class probabilities directly instead of generating label text. On the same backbone and recipe, it incurs a 2.1-point cost in event accuracy for a $1.94\times$ inference speedup (Sec.~\ref{sec:recognition}), which makes 1{,}000-rollout batches practical on a single Graphics Processing Unit (GPU). Task text supplies the objective and plan; recorded outcomes are not recognition inputs. Videos are sampled at two frames per second into stride-one windows of one, two, or four frames.

For each video window $w\in\mathcal W_j$, let $\hat c_{jw}$ be the
highest-probability class over $\{0\}\cup\mathcal D$, with 0 denoting
\emph{none}, and let $p_{jw}$ be the probability of $\hat c_{jw}$.
A window $w$ records a deviation event of factor $i$ (equivalently,
$H_{jwi}=1$) if and only if $\hat c_{jw}=i$ and
$\tau\le p_{jw}<\sigma$ ($\sigma=0.99995$). The lower bound
$\tau=0.95$ keeps only high-confidence events, prioritizing
precision over recall, and the upper bound $\sigma$ excludes
overconfident predictions. All other judgments
are zero. A temporal refinement rule then resolves overlapping f8--f9
events when consolidating the accepted events into final records
(supplementary Sec.~C). A zero record does not establish behavioral absence. This
precision-first screen trades off recall: missed factors reduce
deductions, and the fallback applies only to failures retaining no
scoring factor.

Calibration quantifies this screening through factor--outcome associations estimated from $N=4{,}000$ OpenVLA-OFT rollouts~\cite{kim2025oft} with recorded outcomes. These rollouts are disjoint from the evaluation videos.

For reference rollout $r$ with retained record $X_{ri}$ and outcome $Y_r$,
let $n_i^{\mathrm F}$ and $n_i^{\mathrm S}$ count failed and successful
reference rollouts that record factor $i$, with
$n_i=n_i^{\mathrm F}+n_i^{\mathrm S}$. The screening thresholds of Sec.~\ref{sec:implementation} ensure $a_i>0$, so the penalties in \eqref{eq:factor-costs} are finite. The conditional failure rates and normalized weights are
\begin{equation}\label{eq:factor-weights}
\begin{aligned}
a_i&=\frac{n_i^{\mathrm F}}{n_i}
    =\widehat{\Pr}(Y=0\mid X_i=1),\\
s_i&=\frac{100a_i}{\sum_{h\in\mathcal I}a_h}.
\end{aligned}
\end{equation}
The weights allocate the 100-point budget across factors
($\sum_{i\in\mathcal I}s_i=100$), with $s_i$ the severity of factor $i$
\emph{per recorded occurrence}. Calibration is specific to the reference
policy, task mixture, and recognition protocol (tested in
Sec.~\ref{sec:ablations}).

For each retained record, scoring also determines the deduction under its recorded
outcome. A factor can occur before failure or before successful
recovery; since recovery consumes time without advancing the
task~\cite{feng2026procvlm}, both cases carry positive penalties. Let
$P_i$ and $Q_i$ be the normalized penalties in failed and successful
rollouts, as fractions of the 100-point budget. We set $Q_i=a_iP_i$
(the reference association fixes the success-to-failure ratio) and
require the mean penalty over reference rollouts recording the factor
to equal $s_i/100$, which yields
\begin{equation}\label{eq:factor-costs}
P_i=\frac{s_i}{100a_i(2-a_i)},\qquad Q_i=\frac{s_i}{100(2-a_i)}.
\end{equation}
For every retained factor, $0<Q_i\le P_i$ with $Q_i/P_i=a_i$: successful
recovery preserves a positive deduction. Both structural choices are
validated in Sec.~\ref{sec:ablations}; supplementary
Table~C1 reports the calibration values.

A recorded failure retaining no scoring factor triggers a fallback: the reference mean failure deduction
\begin{equation}\label{eq:fallback-cost}
P_{\mathrm U}=\frac{1}{F}\sum_{i\in\mathcal I}n_i^{\mathrm F}P_i,
\end{equation}
where $F=\sum_r(1-Y_r)$ counts failed reference rollouts. The mean includes co-occurring contributions without assigning a behavioral diagnosis; the calibrated fallback $P_{\mathrm U}\simeq0.8266$ (supplementary Table~C1) deducts about 82.7 points. The fallback assigns deductions to deviations implied by failure but neither detected nor formulated as factors. Failed rollouts therefore incur a deduction despite gaps in detection recall.

For rollout $j$ the fallback trigger is
$U_j=(1-Y_j)\mathbf{1}\!\left[\sum_{i\in\mathcal I}X_{ji}=0\right]$
with $\mathbf1[\cdot]$ the indicator function (a detected event outside $\mathcal I$ does not suppress it), and the total normalized deduction is
\begin{equation}\label{eq:rollout-deduction}
L_j=\sum_{i\in\mathcal I}X_{ji}\bigl[(1-Y_j)P_i+Y_jQ_i\bigr]
    +U_jP_{\mathrm U}.
\end{equation}
For a batch of $M$ rollouts, the score is
\begin{equation}\label{eq:overall-score}
\mathrm{Score}=\max\left\{0,\ 100-\frac{100}{M}\sum_{j=1}^{M}L_j\right\}.
\end{equation}
With $M=1$ the rule scores a single rollout; policy comparisons average
normalized deductions before applying the zero floor.

The outcome-specific contributions are:
\begin{equation}\label{eq:score-contributions}
\begin{gathered}
C_i^{\mathrm F}=\frac{100P_i}{M}\sum_{j=1}^{M}X_{ji}(1-Y_j),\\
C_i^{\mathrm S}=\frac{100Q_i}{M}\sum_{j=1}^{M}X_{ji}Y_j,\qquad
C_{\mathrm U}=\frac{100P_{\mathrm U}}{M}\sum_{j=1}^{M}U_j.
\end{gathered}
\end{equation}
\par

Here $C_i^{\mathrm F}$ and $C_i^{\mathrm S}$ are the failure- and
success-side contributions of factor $i$, and $C_{\mathrm U}$ the
fallback; before flooring, the total deduction is exactly
$\sum_{i\in\mathcal I}(C_i^{\mathrm F}+C_i^{\mathrm S})+C_{\mathrm U}$.
Each contribution combines the fixed penalty and its records, so it
reflects both severity and frequency. The decomposition requires no
assumption of factor independence. Our additive aggregation prioritizes
auditability over maximal agreement. Implicit holistic
scoring~\cite{gunjal2025rubrics} can better match human judgments
within failed executions but lacks recomputable contributions, placing it outside
$\mathcal F_{\mathrm{trace}}$. The
contribution report preserves deductions hidden by zero flooring, and
recomputation is required to audit the score.

\FloatBarrier

\begingroup
\setcounter{topnumber}{5}
\setlength{\dbltextfloatsep}{12pt}
\setlength{\textfloatsep}{6pt plus 2pt minus 1pt}
\setlength{\floatsep}{6pt plus 1pt minus 1pt}
\setlength{\intextsep}{6pt plus 1pt minus 1pt}
\setlength{\abovecaptionskip}{4pt}

\section{Experiments and Analysis}\label{sec:experiments}

\subsection{Evaluation Protocol}\label{sec:experimental-setup}

\textbf{BOT-Events dataset.}
We provide ratings on a 1--5 scale from eight annotators for 800 videos
from ten LIBERO-10 tasks (63\% successes); each video's mean is its
Mean Opinion Score (MOS). We provide 2,394 factor-event annotations for detector
training, from annotators disjoint from the eight quality raters.
Figure~\ref{fig:mos-rater-saturation}(a) shows within-task proportions
of all 6,400 ratings. Mean task-wise absolute-agreement ICC(2,8) is
0.951 (Fig.~\ref{fig:mos-rater-saturation}(c)); pooled eight-rater
corrected split-half reliability is 0.959 across 800 videos
(Fig.~\ref{fig:mos-rater-saturation}(b)). Spearman--Brown extrapolation
from measured single-rater correlation $r_1=0.821$ predicts only a
0.003 gain from a ninth rater, supporting the eight-rater protocol
(supplementary Sec.~A.6.1).

\setcounter{table}{3}
\begin{table*}[!t]
\caption{Per-task Spearman correlation with the 8-rater MOS on 800
videos: (a)~all videos, successes/failures in parentheses;
(b)~successes only. Mean averages the ten tasks;
blue/light blue mark best/second-best per task in (a), gold/light gold
in (b).}
\label{tab:mos-consistency}
\centering
\begingroup
\setlength{\tabcolsep}{2pt}
\begin{minipage}[t]{\linewidth}
\centering\small
\textbf{(a) All videos}\par\smallskip
\small
\renewcommand{\arraystretch}{1.00}
\setlength{\extrarowheight}{0pt}
\begin{tabular*}{\linewidth}{@{\extracolsep{\fill}}lcccccc@{}}
\toprule
Task & $n$ & RoboMeter & PRM-SQS & RoboJev & BOTScorer \\
\midrule
T1 & 68 (51/17) & 0.65 & \cellcolor{tabRankBlueSecond}0.74 & 0.67 & \cellcolor{tabRankBlueBest}\textbf{0.87} \\
T2 & 78 (47/31) & 0.41 & \cellcolor{tabRankBlueSecond}0.81 & 0.66 & \cellcolor{tabRankBlueBest}\textbf{0.88} \\
T3 & 86 (56/30) & 0.57 & \cellcolor{tabRankBlueSecond}0.73 & 0.51 & \cellcolor{tabRankBlueBest}\textbf{0.79} \\
T4 & 66 (66/0) & \llap{$-$}0.06 & \cellcolor{tabRankBlueSecond}0.30 & \llap{$-$}0.24 & \cellcolor{tabRankBlueBest}\textbf{0.33} \\
T5 & 56 (49/7) & 0.25 & 0.40 & \cellcolor{tabRankBlueSecond}0.45 & \cellcolor{tabRankBlueBest}\textbf{0.79} \\
T6 & 138 (57/81) & 0.33 & \cellcolor{tabRankBlueBest}0.79 & 0.63 & \cellcolor{tabRankBlueSecond}\textbf{0.74} \\
T7 & 60 (46/14) & 0.60 & 0.74 & \cellcolor{tabRankBlueSecond}0.83 & \cellcolor{tabRankBlueBest}\textbf{0.90} \\
T8 & 63 (39/24) & 0.63 & \cellcolor{tabRankBlueSecond}0.72 & 0.62 & \cellcolor{tabRankBlueBest}\textbf{0.83} \\
T9 & 111 (53/58) & 0.43 & \cellcolor{tabRankBlueSecond}0.84 & 0.25 & \cellcolor{tabRankBlueBest}\textbf{0.85} \\
T10 & 74 (41/33) & \cellcolor{tabRankBlueSecond}0.72 & 0.71 & 0.46 & \cellcolor{tabRankBlueBest}\textbf{0.90} \\
\midrule
Mean & 800 & 0.45 & \cellcolor{tabRankBlueSecond}0.68 & 0.49 & \cellcolor{tabRankBlueBest}\textbf{0.79} \\
\bottomrule
\end{tabular*}
\end{minipage}\par\medskip
\begin{minipage}[t]{\linewidth}
\centering\small
\textbf{(b) Successes only}\par\smallskip
\small
\renewcommand{\arraystretch}{1.00}
\setlength{\extrarowheight}{0pt}
\begin{tabular*}{\linewidth}{@{\extracolsep{\fill}}lcccccc@{}}
\toprule
Task & $n$ & RoboMeter & PRM-SQS & RoboJev & BOTScorer \\
\midrule
T1 & 51 & 0.27 & \cellcolor{tabOursGoldSecond}0.39 & 0.27 & \cellcolor{tabOursGold}\textbf{0.70} \\
T2 & 47 & \llap{$-$}0.10 & \cellcolor{tabOursGoldSecond}0.32 & 0.04 & \cellcolor{tabOursGold}\textbf{0.69} \\
T3 & 56 & \llap{$-$}0.15 & \cellcolor{tabOursGoldSecond}0.14 & \llap{$-$}0.10 & \cellcolor{tabOursGold}\textbf{0.25} \\
T4 & 66 & \llap{$-$}0.06 & \cellcolor{tabOursGoldSecond}0.30 & \llap{$-$}0.24 & \cellcolor{tabOursGold}\textbf{0.33} \\
T5 & 49 & \llap{$-$}0.12 & 0.11 & \cellcolor{tabOursGoldSecond}0.18 & \cellcolor{tabOursGold}\textbf{0.68} \\
T6 & 57 & 0.32 & \cellcolor{tabOursGold}0.38 & 0.09 & \cellcolor{tabOursGoldSecond}\textbf{0.37} \\
T7 & 46 & 0.17 & 0.46 & \cellcolor{tabOursGoldSecond}0.64 & \cellcolor{tabOursGold}\textbf{0.79} \\
T8 & 39 & \cellcolor{tabOursGoldSecond}0.10 & \llap{$-$}0.13 & 0.02 & \cellcolor{tabOursGold}\textbf{0.51} \\
T9 & 53 & \llap{$-$}0.01 & \cellcolor{tabOursGoldSecond}0.29 & 0.11 & \cellcolor{tabOursGold}\textbf{0.82} \\
T10 & 41 & \cellcolor{tabOursGoldSecond}0.12 & \llap{$-$}0.08 & \llap{$-$}0.19 & \cellcolor{tabOursGold}\textbf{0.77} \\
\midrule
Mean & 505 & 0.06 & \cellcolor{tabOursGoldSecond}0.22 & 0.08 & \cellcolor{tabOursGold}\textbf{0.59} \\
\bottomrule
\end{tabular*}
\end{minipage}
\endgroup
\end{table*}

\setcounter{table}{2}
\begin{figuretablepair}
\begin{minipage}{\linewidth}
\setlength{\intextsep}{0pt}
\begin{table}[H]
\caption{Event detection on 239 test windows. Acc.: accuracy (\%);
mF1: macro-F1; s/rollout: inference time. Blue/light blue: best/second
baseline; gold bold: ours. SFT rows decode labels generatively; ours
returns class probabilities via a discriminative head. All timings on a single NVIDIA GeForce
RTX 3090 (24\,GB).}
\label{tab:detector}
\centering\small
\setlength{\tabcolsep}{2.0pt}
\small
\renewcommand{\arraystretch}{1.00}
\setlength{\extrarowheight}{0pt}
\begin{tabular*}{\linewidth}{@{\extracolsep{\fill}}lccc@{}}
\toprule
Model & Acc. (\%) & mF1 & s/rollout \\
\midrule
Qwen3-VL-2B \emph{(Zero-shot)} & 7.11 & 0.031 & 6.2 \\
InternVL3.5-1B \emph{(Zero-shot)} & 0.84 & 0.036 & 7.6 \\
Qwen3-VL-2B \emph{(SFT)} & \cellcolor{tabRankBlueBest}92.47 & \cellcolor{tabRankBlueBest}0.888 & 6.2 \\
Qwen3.5-2B \emph{(SFT)} & 86.61 & 0.814 & 13.6 \\
Qwen3.5-0.8B \emph{(SFT)} & 89.54 & \cellcolor{tabRankBlueSecond}0.876 & 12.7 \\
InternVL3-2B \emph{(SFT)} & 89.54 & 0.870 & 6.7 \\
InternVL3-1B \emph{(SFT)} & 87.45 & 0.848 & \cellcolor{tabRankBlueBest}4.7 \\
InternVL3.5-1B \emph{(SFT)} & \cellcolor{tabRankBlueSecond}90.79 & 0.853 & \cellcolor{tabRankBlueSecond}6.1 \\
\midrule
Factor-SFT-Qwen (ours) & \cellcolor{tabOursGold}\textbf{90.38} & \cellcolor{tabOursGold}\textbf{0.873} & \cellcolor{tabOursGold}\textbf{3.2} \\
\bottomrule
\end{tabular*}
\end{table}
\end{minipage}
\end{figuretablepair}
\setcounter{table}{4}

The 2,394 factor-event examples are split by rollout into 1,916 training,
239 validation, and 239 test examples (training details: supplementary
Sec.~A.1). Reference analyses use 4,000
OpenVLA-OFT rollouts disjoint from the 800 evaluation videos, with
factors, recognition rules, and penalties fixed. All experiments,
including detector training and timing in Table~\ref{tab:detector},
use one NVIDIA GeForce RTX 3090 (24\,GB). RoboMeter, PRM-SQS, and RoboJev use released models
and official protocols without human ratings; PRM-SQS applies
PRM-as-a-Judge 1.5 to GRM-2.0-8B progress curves from Robo-Dopamine
(CVPR 2026)~\cite{tan2026robodopamine}. All three baselines use the same
800 videos and task splits. Factor-SFT-Qwen receives Supervised
Fine-Tuning (SFT) on 1,916 factor-labeled windows; baselines run
zero-shot, so supervision is unmatched. Recorded outcomes $Y_j$ enter
only the scoring rule (Sec.~\ref{sec:score-aggregation}), ablated in
Sec.~\ref{sec:ablations} (supplementary Sec.~A.6).
Eval-Actions/AutoEval~\cite{liu2026evalactions} has no released trained
checkpoints or external-rollout inference pipeline; we compare it
analytically in Sec.~\ref{sec:related}.

\subsection{Event Detection}
\label{sec:recognition}
Table~\ref{tab:detector} compares backbones and adaptation settings,
including generative SFT on the same Qwen3-VL-2B backbone and recipe
(supplementary Table~E1).
Factor-SFT-Qwen reaches 90.38\% event accuracy and 0.873 macro-F1
across the seven trained factors and \emph{none}.

\textbf{A $1.94\times$ speedup at a 2.1-point accuracy cost.}
Inference falls from 6.2 to 3.2 s per rollout, and the full pipeline
takes about 4.8 s (supplementary Sec.~A.1).

\textbf{Recognition within successful videos.}
In the audit of 62 flawed and 18 clean successful OpenVLA-OFT videos (a
success is flawed if annotators identify at least one factor in
$\mathcal I$), Factor-SFT-Qwen correctly classifies 52 (83.9\%) and 15 (83.3\%),
respectively. No zero-shot setting matches both rates
(supplementary Table~A1). Visual perturbations reduce
the frozen detector's accuracy; quantitative comparisons use common
observation conditions (supplementary Sec.~A.4).

\begin{figure}[!t]
\centering
\includegraphics[width=0.89\linewidth]{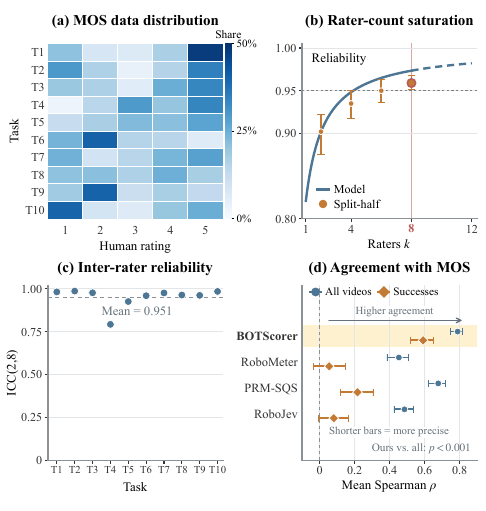}
\caption{Human ratings and agreement with MOS on 800 videos.
(a) Within-task percentages of 6,400 individual ratings; rows sum to 100\%.
(b) Model and corrected split-half reliability by rater count;
bars: enumerated min--max; dashed: extrapolation beyond eight raters.
(c) Task-wise absolute-agreement ICC(2,8); dashed: ten-task mean.
(d) Mean per-task Spearman correlations and 95\% CIs from 2,000
paired within-task video resamples. Blue: all videos; orange: successes;
shading: BOTScorer.}
\label{fig:mos-rater-saturation}
\end{figure}

\subsection{Quality Assessment of BOT-Scale}
\label{sec:mos-consistency}
\label{sec:reference-robustness}

Table~\ref{tab:mos-consistency} reports per-task Spearman correlations
with the MOS. BOTScorer is highest on nine of ten tasks over all videos
(mean $\rho=0.789$) and leads the harder successes-only setting, which
tests differences among executions that all succeed. BOTScorer reaches a mean
$\rho=0.592$, while every baseline stays at or below 0.22 (RoboJev and RoboMeter
saturate on successes, $\rho=0.081$ and $0.055$; their pooled
correlations reflect success--failure differences). Pooled correlations
rank all 800 videos together; per-task means average the within-task
correlations. These two quantities are not directly comparable; supplementary analyses use
the pooled form (Table~D1). BOTScorer distinguishes successful executions through
factor records.

Paired within-task bootstrap ($B=2{,}000$, ten fixed tasks) gives
95\% Confidence Intervals (CIs) of $[0.749,0.817]$ (all videos) and $[0.518,0.649]$
(successes) for BOTScorer's mean $\rho$; all six paired mean-difference
CIs are positive ($p<0.001$; Fig.~\ref{fig:mos-rater-saturation}(d)).
Rating reliability is detailed in supplementary
Sec.~A.6. Aggregation and outcome conditioning are
tested in Sec.~\ref{sec:ablations}; within the 505 successes, ranking
is label-free.

\textbf{Reference support.}
Calibration stabilizes with reference-set size; refined records have
limited overlap (maximum $|\varphi|=0.338$; Fig.~\ref{fig:factor-outcome};
supplementary Sec.~A.3). The additive decomposition
requires no independence assumption despite cross-layer nesting. All 40 f6 records
come from failures and co-occur with f12; removing f6 leaves successes-only
agreement unchanged (Sec.~\ref{sec:ablations}). It supports failure
attribution, with less certainty between f6 and f12 than in the total.

\subsection{Ablations}
\label{sec:ablations}

The scoring rule is layer-agnostic: layers never enter
Eqs.~\ref{eq:rollout-deduction}--\ref{eq:score-contributions}. We ablate
factor sets, penalties, and calibration references by recalibrating and
rescoring frozen records (Table~\ref{tab:ablations}; supplementary
Sec.~D). We also compare aggregation rules
on the same records, including Gradient-Boosted Decision Trees (GBDT)
and a Multilayer Perceptron (MLP). Learned aggregators use task-level
five-fold Cross-Validation (CV).

\textbf{Factor set.}
Removing f6 or f7 improves full-pool per-task agreement
(Table~\ref{tab:ablations}a), while six of seven removals reduce pooled
successes-only agreement; f6 is the exception. Across all 127 subsets,
the deployed set, chosen without rating supervision, is best on pooled
successes-only agreement. No subset dominates all axes; the subset
search is in-sample (supplementary
Table~D1).

\textbf{Penalty structure.}
Unified penalties lower mean per-task agreement and raise the
failure-pool mean score from 20.5 to 60.2 (Table~\ref{tab:ablations}b).
Since $P_i/Q_i=1/a_i$, a unified penalty under-penalizes one outcome
side. Equal weights improve the full-pool per-task mean but reduce
pooled and pooled successes-only agreement (supplementary
Table~D3). The deployed scheme balances these agreement
axes while preserving outcome-specific penalties.

\textls[-10]{\textbf{Aggregation under matched records.}
On fixed detector records from 800 videos, we evaluate MOS-supervised
Ridge, GBDT and MLP by task-level five-fold cross-validation. All arms
share seven factor indicators, a constant outcome and no fallback
on 505 successes. The frozen-penalty rule leads despite using no MOS-trained weights. The
learned arms fit ratings and fall outside $\mathcal F_{\mathrm{trace}}$
(Table~\ref{tab:ablations}c; supplementary
Sec.~A.7).}\begin{table}[!t]
\caption{Scoring ablations on 800 videos. Bold: key results. M10: mean per-task
Spearman; $\Delta$ in (a): change from full-7, computed from
unrounded values. S1/S2: per-factor
weights, split/unified penalties; S3/S4: equal weights, split/unified
(S1 deployed). Fail: mean score on 295 failures; Flips: rank reversals
vs. S1 (28 pairs). $^{**}$/$^{***}$: $p\le0.05/0.01$ (paired bootstrap,
$B{=}2{,}000$). (c) Pooled Spearman on 505 successes; learned arms use
MOS-supervised task-level five-fold CV. (d) $\sigma$: logit-noise scale;
M10-S: successes-only M10. Medians and mean Kendall $\bar\tau_K$ over
2,000 draws per nonzero level; 0 is the unperturbed baseline.}
\label{tab:ablations}
\centering\small
\renewcommand{\arraystretch}{1.00}
\setlength{\extrarowheight}{0pt}
\setlength{\tabcolsep}{1.5pt}
\begin{minipage}[t]{0.70\linewidth}
\centering
\textbf{(a) Leave-one-out}\par\smallskip
\begin{tabular}{@{}lcc@{\hspace{4pt}}lcc@{}}
\toprule
Set & M10 & $\Delta$ & Set & M10 & $\Delta$ \\
\midrule
full-7 & 0.79 & --- & $-$f7 & 0.85 & $\mathbf{+0.06}^{***}$ \\
$-$f1 & 0.79 & $0.00$ & $-$f8 & 0.78 & $-0.01^{***}$ \\
$-$f4 & 0.78 & $-0.01^{**}$ & $-$f9 & 0.78 & $-0.01^{**}$ \\
$-$f6 & 0.80 & $\mathbf{+0.02}^{***}$ & $-$f12 & 0.78 & $-0.01$ \\
\bottomrule
\end{tabular}
\end{minipage}\hfill
\begin{minipage}[t]{0.28\linewidth}
\centering
\textbf{(b) Penalties}\par\smallskip
\setlength{\tabcolsep}{1pt}
\begin{tabular}{@{}lccc@{}}
\toprule
S & M10 & Fail & Flips \\
\midrule
S1 & 0.79 & \textbf{20.5} & 0 \\
S2 & 0.74 & \textbf{60.2} & 2 \\
S3 & 0.80 & 17.9 & 2 \\
S4 & 0.75 & 39.1 & 6 \\
\bottomrule
\end{tabular}
\end{minipage}
\par\smallskip
\begin{minipage}[t]{0.45\linewidth}
\centering
\textbf{(c) Aggregation}\par\smallskip
\begin{tabular*}{\linewidth}{@{\extracolsep{\fill}}lc@{}}
\toprule
Aggregator & Pooled $\rho$ \\
\midrule
Ours (frozen) & \textbf{0.741} \\
Ridge & 0.554 \\
GBDT & 0.563 \\
MLP & 0.511 \\
\bottomrule
\end{tabular*}
\end{minipage}\hfill
\begin{minipage}[t]{0.53\linewidth}
\centering
\textbf{(d) Detection noise}\par\smallskip
\setlength{\tabcolsep}{1pt}
\begin{tabular*}{\linewidth}{@{\extracolsep{\fill}}cccc@{}}
\toprule
$\sigma$ & M10 & M10-S & $\bar\tau_K$ \\
\midrule
0 & 0.789 & 0.592 & 1.000 \\
0.21 & 0.787 & 0.585 & 0.992 \\
0.42 & 0.785 & 0.575 & 0.987 \\
1.05 & \textbf{0.775} & 0.539 & \textbf{0.976} \\
\bottomrule
\end{tabular*}
\end{minipage}
\end{table}

\begin{figuretablepair}
\begin{minipage}{\linewidth}
\setlength{\intextsep}{0pt}
\begin{table}[H]
\caption{Selected VLA scores with fixed penalties (1,000 rollouts/model;
100/task). SR: success rate (\%); Score: $100-$total deduction.
Success: deductions from successes; f7--f9: per-factor deductions
from both outcomes. All deductions: points/100 rollouts.
Ranks over all eight models; red bold marks the highest value per
column---best for Score, largest (worst) deduction for Success and
f7--f9. Full results:
supplementary Table~A8. Values rounded.}
\label{tab:main}
\centering\small
\setlength{\tabcolsep}{2pt}
\small
\renewcommand{\arraystretch}{1.00}
\setlength{\extrarowheight}{0pt}
\begin{tabular*}{\linewidth}{@{\extracolsep{\fill}}lcccccc@{}}
\toprule
Model & Score $\uparrow$ & SR & Success & f7 & f8 & f9 \\
\midrule
StarVLA-OFT & \textcolor{blue}{93.52 (2)} & \textcolor{blue}{94.2 (3)} & 2.87 & 0.92 & 0.67 & 1.39 \\
OpenVLA-OFT & \textcolor{red}{92.95 (3)} & \textcolor{red}{94.0 (4)} & 3.36 & 0.98 & 0.97 & 1.31 \\
SimVLA & 92.58 (4) & 94.3 (2) & \textcolor{red}{\textbf{4.33}} & 1.28 & 0.71 & 1.09 \\
$\pi$0.5 & 90.41 (6) & 91.4 (6) & 3.55 & 1.04 & \textcolor{red}{\textbf{1.44}} & 1.88 \\
UniVLA & 90.31 (7) & 90.5 (7) & 3.53 & 1.00 & 0.92 & 1.83 \\
\bottomrule
\end{tabular*}
\end{table}
\vspace{2pt}
\begin{figure}[H]
\centering
\includegraphics[width=\linewidth]{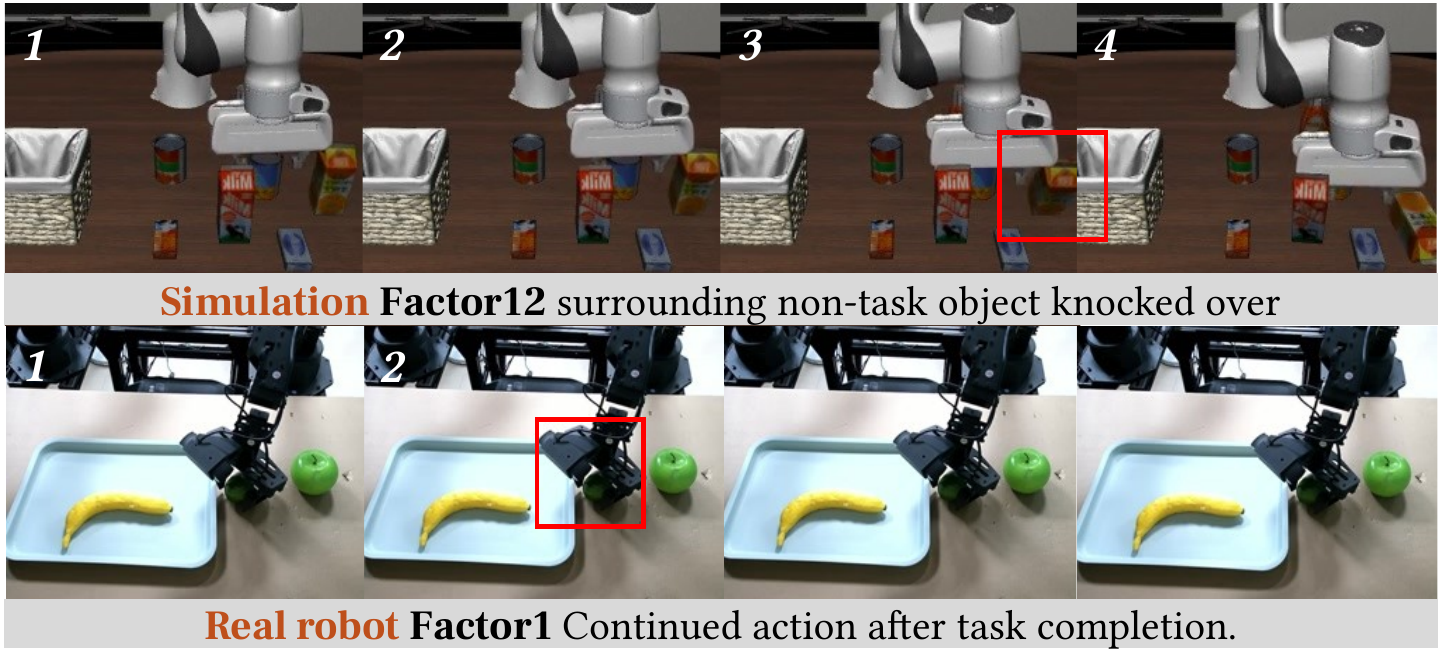}
\caption{\textbf{Simulation and real robot.} Successful rollouts with
detected deviation events boxed. Top: f12, a non-task object knocked over.
Bottom: f1, continued action after task completion.}
\label{fig:libero-records}
\end{figure}
\end{minipage}
\end{figuretablepair}

\begingroup\clubpenalty=10000
\textbf{Cross-policy calibration.}
Leave-one-policy-out recalibration preserves the eight-model ranking
(Kendall $\tau=1.0$), with pooled successes-only correlation changing by
$-0.093$ (supplementary Table~D5). We recalibrate outcome
associations from recorded outcomes without human
ratings; the tested policies share a benchmark and are partly related.\par\endgroup

\textbf{Detection-probability sensitivity.}
We add Gaussian noise in logit space to archived probabilities
($p\ge0.85$; 2,000 draws/level), with penalties and f8--f9 ordering fixed.
At the largest noise level, median M10 is 0.775 (95\% perturbation band
$[0.760,0.789]$); rankings change by at most one near-tie pair
(Table~\ref{tab:ablations}d). Supplementary
Sec.~D.5 reports the full score-sensitivity profiles under this model.

\subsection{Application to VLA Model Evaluation}\label{sec:model-interpretation}

\setcounter{topnumber}{5}
\setlength{\textfloatsep}{2pt plus 1pt minus 1pt}

Each of the eight released checkpoints~\cite{xiaomi2026checkpoint,starvla2026code,kim2025oft,luo2026simvlacode,xu2025vlacache,intelligence2025pi05,bu2025univla,nvidia2026grootn17}
provides 1,000 rollouts (100/task), disjoint from the 4,000-rollout
reference set; only the factor detector is
trained, and all batches use the same frozen penalties. Table~\ref{tab:main}
shows five representative models. Similar
success rates can mask quality differences: StarVLA-OFT and
OpenVLA-OFT differ by only 0.2 SR but 0.57 score points, with 0.50
from successful rollouts. SimVLA reaches 94.3\% SR but incurs 4.33
success-side deduction points versus StarVLA-OFT's 2.87, reflecting
flawed successes that SR does not penalize. Per-factor contributions
distinguish these near-tied models (Fig.~\ref{fig:libero-records}, top).

\begingroup
\textbf{Real-robot demonstration.}
With detector and simulation penalties unchanged, BOTScorer produces
records and scores for three successful OpenVLA-OFT rollouts on ALOHA
(Fig.~\ref{fig:libero-records}, bottom). The demonstration is qualitative;
quantitative claims remain restricted to simulation (supplementary
Sec.~A.8).
\par\endgroup

\textbf{Boundaries.}
With 1,000 rollouts per model, 2,000 bootstrap resamples give 95\% CI
half-widths of 0.9--1.5 points. Adjacent gaps below two points,
including the StarVLA-OFT--OpenVLA-OFT pair, do not establish rank
order; middle ranks are ties within sampling uncertainty. For such
near-tied pairs, the per-factor contributions still decompose the
observed gap exactly and recomputably; this attributes the difference
without ranking the two policies. Only the UniVLA--GR00T~N1.7 gap is
significant ($p=0.048$; supplementary
Table~A8).

\endgroup

\section{Conclusion}\label{sec:conclusion}
We introduced BOT-Scale, a traceable evaluation methodology that adapts
clinical assessment principles to robotic manipulation. It computes
scores and explanations together from retained behavioral-factor
records and outcome-conditioned penalties. BOTScorer implements the
method, and BOT-Events supports training and evaluation. Experiments
in single-arm simulation show agreement with human quality judgments
and reveal execution differences that success rates alone miss. The
resulting records can guide targeted improvements in robotic
manipulation. Future work will refine the detector and extend
task-specific factor sets to other embodiments and benchmarks.

\FloatBarrier
\section*{LLM Usage Statement}
We used ChatGPT and Claude for language polishing, grammar checking, assistance with part of data preprocessing and coding, and obtaining feedback on the manuscript. All AI-assisted outputs were reviewed and verified by the authors.

\end{document}